\documentclass[runningheads]{llncs}

\usepackage[T1]{fontenc}

\usepackage{graphicx}
\usepackage{amsmath,amssymb} 
\usepackage{lineno}
\usepackage{color}
\usepackage{bussproofs}

\usepackage{times}
\usepackage{epsfig}
\usepackage{eufrak}
\usepackage{hyperref}

\DeclareMathOperator{\Path}{Path}

\DeclareMathOperator{\edge}{edge}

\DeclareMathOperator{\connect}{connect}

\DeclareMathOperator{\Vect}{Vec}

\DeclareMathOperator{\fix}{fix}

\DeclareMathOperator{\nul}{null}
\DeclareMathOperator{\trans}{transform}
\DeclareMathOperator{\Eq}{Eq}

\DeclareMathOperator{\Type}{Type}

\DeclareMathOperator{\tuple}{tuple}
\DeclareMathOperator{\funapp}{fun-app}
\DeclareMathOperator{\rewrite}{rewrite}
\DeclareMathOperator{\update}{update}

\DeclareMathOperator{\isGLTSBisim}{isGLTSBisim}
\DeclareMathOperator{\is2GLTSBisim}{is2GLTSBisim}

\DeclareMathOperator{\random}{random}
\DeclareMathOperator{\sample}{sample}
\DeclareMathOperator{\thunk}{thunk}
\DeclareMathOperator{\Distribution}{Distribution}
\DeclareMathOperator{\match}{match}

\allowdisplaybreaks

\begin{document}



\title{A meta-probabilistic-programming language for bisimulation of probabilistic and non-well-founded type systems} 




\author{Jonathan Warrell$^{1,2}$, Alexey Potapov$^2$, Adam Vandervorst$^2$, Ben Goertzel$^2$}

\authorrunning{J. Warrell, A. Potapov, A. Vandervorst, B. Goertzel}
\titlerunning{Meta-probabilistic-programming and bisimulation}

\institute{\em $^1$Yale University, $^2$SingularityNET}
\maketitle

\begin{abstract}
We introduce a formal meta-language for probabilistic programming, capable of expressing both programs and the type systems in which they are embedded.  We are motivated here by the desire to allow an AGI to learn not only relevant knowledge (programs/proofs), but also appropriate ways of reasoning (logics/type systems).  We draw on the frameworks of cubical type theory and dependent typed metagraphs to formalize our approach.  In doing so, we show that specific constructions within the meta-language can be related via bisimulation (implying path equivalence) to the type systems they correspond.  This allows our approach to provide a convenient means of deriving synthetic denotational semantics for various type systems.  Particularly, we derive bisimulations for pure type systems (PTS), and probabilistic dependent type systems (PDTS).  We discuss further the relationship of PTS to non-well-founded set theory, and demonstrate the feasibility of our approach with an implementation of a bisimulation proof in a Guarded Cubical Type Theory type checker.
\end{abstract}

\section{Introduction}\label{sec:intro}
Probabilistic programming offers a fertile ground between logic-based and machine-learning-based approaches to A(G)I.  Formalization within type theory offers a rigorous approach to deriving semantics for probabilistic languages \cite{staton_16}, and formalization of dependently typed probabilistic languages offers the promise of drawing a tight connection with probabilistic logics of various kinds (e.g. Markov Logic \cite{warrell_18}, Probabilistic Paraconsistent Logic \cite{goertzel_20b}).

While the exploration of such individual systems is highly important, we might consider more abstractly how to embody general principles for the formation of diverse probabilistic type systems, logics, and programming languages within a single meta-language.  Such a language can be considered a meta-theoretical language or logical framework for expressing individual type systems and logics.  However, previous frameworks (such as \cite{harper_12}) have not been designed with probabilistic type systems and logics specifically in mind.   Here, we outline a formal language, $\mathbb{M}$, designed for such a purpose.  This language is intended as a formal model of the MeTTa language, currently being developed as part of the OpenCog project \cite{potapov_21,goertzel_21,trueagi_21}.  The language allows for (probabilistic) reasoning not only about the knowledge embedded in a system, but also about the logic employed by the system itself.

Our approach may also be seen in relation to recent methods to derive synthetic denotational semantics for logical systems using guarded cubical type theory (GCTT) \cite{vezzosi_21,mogelborg_19b}.  Such approaches are particularly promising, offering as they do a unified approach to deriving semantics for recursive datatypes as final co-algebras of appropriate functors in the context of a formulation of univalent type theory with a fully computational semantics.  We draw on methods from \cite{mogelborg_19} to formalize our approach in this context.  This allows us to rigorously define the relationship between an object-language and its expression in our meta-language as one of bisimulation, corresponding to path equivalence in GCTT.  We further show how dependently typed metagraphs can be formalized in GCTT as the basis for our framework \cite{goertzel_20,mokhov_17}, and how this leads to systems embedding natural type-theoretic equivalents of non-well-founded sets.

We begin by developing a general framework for representing metagraphs in GCTT, before outlining how the final co-algebra of a labeled transition system over this recursive datatype can be used to model our meta-language.  We then derive bisimulations for various object-languages in our system, including simply typed (and untyped) lambda calulus, pure type systems, and probabilistic dependent type systems, hence deriving synthetic denotational semantics for these systems.  Finally, we demonstrate the feasibility of our approach with an implementation of a bisimulation proof for a small-scale type system in a Guarded Cubical Type Theory type checker \cite{birkedal_16}, before concluding with a discussion.

\section{Labeled metagraphs as a guarded recursive datatype}\label{sec:3}

We begin by defining a recursive datatype for typed metagraphs ($\mathcal{M}_{(\mathcal{T},\mathcal{L},\preceq_T)}$) using guarded cubical type theory.  Here, $\mathcal{T},\mathcal{L}$ are types of type-symbols and edge labels respectively, and $\preceq_T:\mathcal{T}\times \mathcal{T} \rightarrow \mathbb{B}$ is a partial order on type-symbols.  The recursive datatype is defined as the final co-algebra of the functor $\mathcal{M}'_{(\mathcal{T},\mathcal{L},\preceq_T)}(A)$, which when applied to type $A$ returns the following datatype (letting $\Delta$ stand for the assumptions $\mathcal{L}, \mathcal{T}, A:\mathcal{U}_0$; the $\epsilon,\edge,$ and $\connect$ constructors used here follow the approach of \cite{mokhov_17} and \cite{goertzel_20}):

\begin{prooftree}
\AxiomC{$\Gamma \vdash \Delta$}
\UnaryInfC{$\Gamma \vdash \mathcal{M}'_{(\mathcal{T},\mathcal{L})}(A)$}
\end{prooftree}
\begin{prooftree}
\AxiomC{$\Gamma \vdash \Delta$}
\UnaryInfC{$\Gamma \vdash \epsilon:\mathcal{M}'_{(\mathcal{T},\mathcal{L})}(A)$}
\end{prooftree}
\begin{prooftree}
\AxiomC{$\Gamma \vdash \Delta, n:\mathbb{N}, t_0:\mathcal{T}, t:\Vect(n,\mathcal{T}), l_0:\mathcal{L}$}
\UnaryInfC{$\Gamma \vdash \edge(n,t_0,l_0,t):\mathcal{M}'_{(\mathcal{T},\mathcal{L})}(A)$}
\end{prooftree}
\begin{prooftree}
\AxiomC{$\Gamma \vdash \Delta, a_1, a_2:A, t_0:\mathcal{T}, l_0:\mathcal{L}, q:\mathbb{N}\rightarrow\mathbb{N}_{0,\infty}$}
\UnaryInfC{$\Gamma \vdash \connect(a_1,a_2,t_0,l_0,q):\mathcal{M}'_{(\mathcal{T},\mathcal{L})}(A)$}
\end{prooftree}
\begin{eqnarray}\label{eq:1}
\end{eqnarray}

\noindent where $\Vect(n,A)$ is the type of vectors over $A$ of length $n$, and $\mathbb{N}_{0,\infty}$ is $\mathbb{N}$ extended with $0$ and $\infty$.  We note that for notational convenience, we do not explicitly include target labels/indices in the definition of $\mathcal{M}'_{(\mathcal{T},\mathcal{L})}(A)$ above (in contrast to \cite{goertzel_20}, where $\mathcal{L}$ refers to target indices and $\mathcal{V}$ is used for edge values).  If explicit indices are required to identify target 'levels', these may be included by letting $\mathcal{L}=\mathcal{L}_0 \times \sum_n \Vect(n,\mathbb{N})$, so that each edge label is paired with a vector of target indices.  $\mathcal{M}_{(\mathcal{T},\mathcal{L},\preceq_T)}$ is then defined as a final fixed-point of $\mathcal{M}'_{(\mathcal{T},\mathcal{L})}$, such that a set of constraints are satisfied:
\begin{eqnarray}
\mathcal{M}_{(\mathcal{T},\mathcal{L},\preceq_T)} &=& \sum M:\nu(\mathcal{M}'_{(\mathcal{T},\mathcal{L})}).C(M,\preceq_T)
\end{eqnarray}
\noindent where $C(M,\preceq_T)$ represents the constraints:
\begin{eqnarray}
C(M,\preceq_T) &=& \forall n_1,n_2:\mathbb{N}, t_1,t_2:\mathcal{T} \nonumber \\
&& f(M,n_1)=t_1 \wedge \nonumber \\
&& f(M,n_2)=t_2 \wedge \nonumber \\
&& q'_M(n_1)=n_2 \Rightarrow t_1 \preceq_T t_2
\end{eqnarray}
\noindent  Here, $f(M,n)$ represents a function, which for metagraph $M$ returns the type of its $n$'th edge or target.  Specifically, when $M$ is of the form $\edge(n,t_0,l_0,t)$, $f(M,0)$ is the type of the edge, and $f(M,n>0)$ is the type of the $n$'th target, and when $M$ is of the form $\connect(a_1,a_2,t_0,l_0,q)$, $f(M,0)$ is the type of the whole metagraph, while the types of the edges/targets of $a_1$ and $a_2$ are interleaved when evaluating $f(M,n>0)$ for odd/even values of $n$ respectively.  Further, the function $q'_M:\mathbb{N}\rightarrow\mathbb{N}_{0,\infty}$ is recursively defined on $\nu(\mathcal{M}'_{(\mathcal{T},\mathcal{L})})$ (via the $q$ function in the $\connect$ constructor of Eq. \ref{eq:1}) to indicate that the $n_1$'th target of $M$ is connected to the $n_2$'th edge/target of $M$, whenever $q(n_1)=n_2$, with $n_2=\infty$ indicating that the target has no connection.  $C(M,\preceq_T)$ thus provides a set of constraints that ensure the connections in a metagraph respect the $\preceq_T$ relation; further constraints are needed to ensure for instance that targets receive input from only one other target (as may be appropriate for some metagraphs).  Further,  $\nu=\fix X.F(\triangleright(\alpha : \mathbb{T}).X[\alpha]))$ is the guarded fixed-point operator \cite{mogelborg_19}.  By \cite{mogelborg_19}, Prop. 3.2, $\mathcal{M}_{(\mathcal{T},\mathcal{L},\preceq_T)}$ is both a subset of the initial algebra and final coalgebra of $\mathcal{M}'_{(\mathcal{T},\mathcal{L})}\circ\triangleright$.  Finally, we note that our $\connect$ constructor corresponds to $\text{Connect}_Q$ in \cite{goertzel_20}, and the $\text{Union}$ constructor is simply $\connect$ with $q(n)=\infty$ for all $n$ (meaning that no new connections are added).

\begin{figure*}[!t]
\centering
\includegraphics[width=3in]{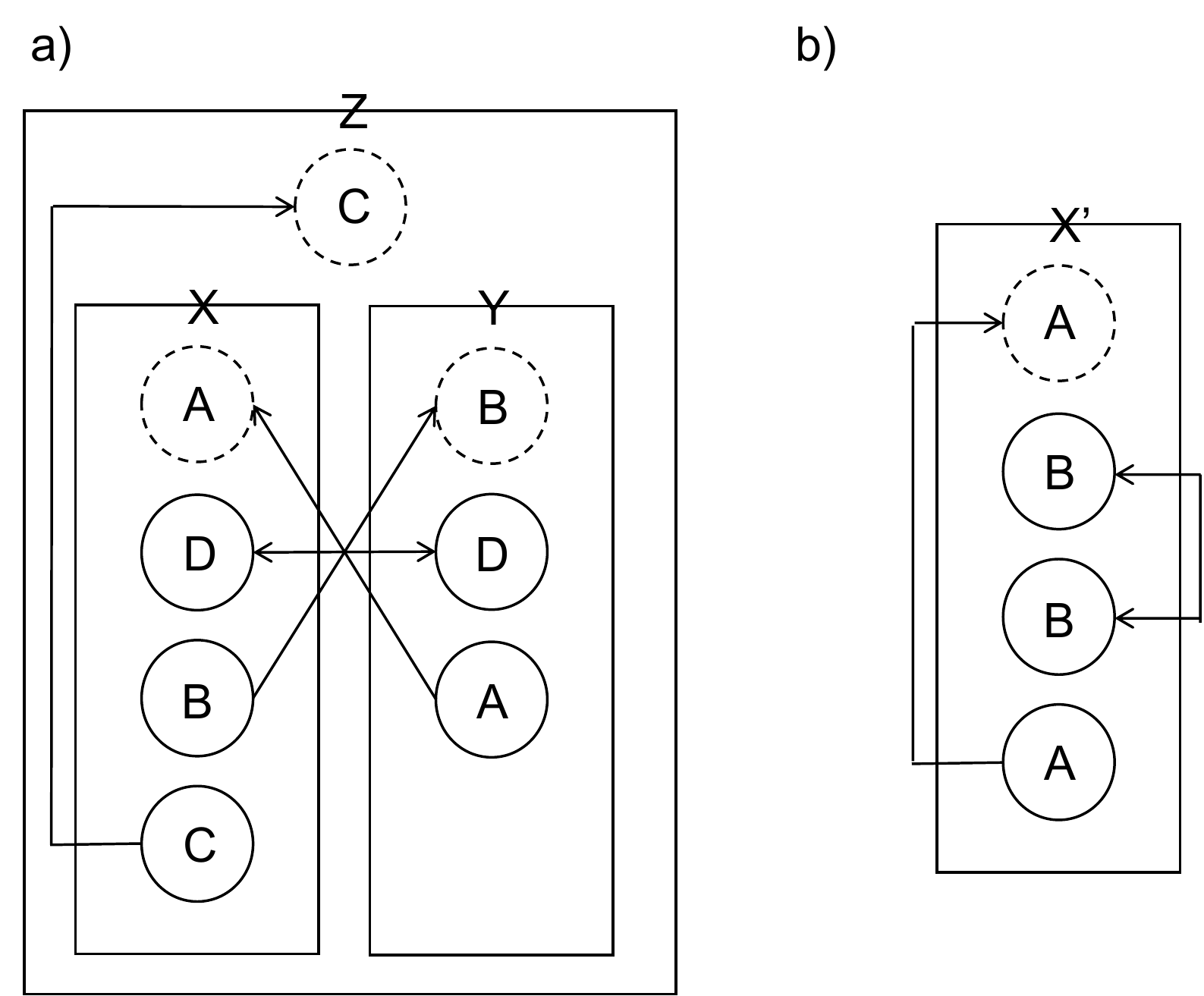}
\caption{Typed metagraph examples.  Boxes show metagraphs, which may be single edges (containing no further boxes) or include several edges.  Solid circles edge target types and dotted circles show metagraph types.  Arrows show target-target or target-edge connections.  Metagraph letter names are shown on the box of the metagraph to which they refer in the text.}
\label{fig:fig1}
\end{figure*}

We briefly give some examples of typed metagraphs.  For convenience, we set $\mathcal{L}=\{\nul\}$, and $\mathcal{T}=\{A,B,C,D,\top\}$, with $\preceq_T$ the identity relation along with $t\preceq_T \top$ for all $t$.  In our first example, we can construct metagraphs $X = \edge(3,A,\nul,[D,B,C])$, and $Y = \edge(2,B,\nul,[D,A])$.  Then, a combined graph can be constructed as $Z'=\connect(X,Y,\top,\nul,\{(1,1),(2,0)\})$, $Z''=\connect(Y,X,\top,\nul,\{(1,1),(2,0)\})$, $Z'''=\connect(Z',Z'',\top,\nul,\{\})$, $Z=\connect(X,Z''',C,\nul,\{(3,0)\})$.  The entire metagraph is shown in Fig. \ref{fig:fig1}A.  We note that, in general, any metagraph with a finite number of edges and targets can be represented by a term in the initial algebra of $\mathcal{M}'_{(\mathcal{T},\mathcal{L})}$ (as is $Z$).  Some graphs, however, may be conveniently be represented also by terms in the final coalgebra.  Consider for instance Fig. \ref{fig:fig1}B.  Here, we may define $X''=\edge(3,\top,\nul,[B,B,A])$ and $X'=\connect(X'',X'',A,\nul,\{(1,2),(2,1),(3,0)\})$, representing $X'$ by a term in the initial algebra (suppressing visualization of the $X''$ subgraph).  Alternatively, we may define $X'_{co}=\connect(\edge(3,A,\nul,[B,B,A]),\allowbreak X'_{co},A,\nul,\{(1,2),(2,1),(3,0)\})$, which implicitly determines a term in the coalgebra as a solution to the recursive equation.

\section{$\mathbb{M}$ as the final coalgebra of a labeled transition system}\label{sec:3}

We define the formal meta-probabilistic-programming language, $\mathbb{M}$, as a labeled transition system over typed metagraphs.  Here, we are interested in typed metagraphs with a particular form.  Specifically, we begin by defining $\mathcal{T} $ by the abstract syntax:
\begin{eqnarray}\label{eq:metta1}
\mathcal{T} &::=& t_n \;\;|\;\; \mathcal{T} \rightarrow \mathcal{T} \;\;|\;\; \prod a:\mathcal{T}.\mathcal{M}_{\mathbb{M}} \;\;|\;\;  \nonumber \\
&& \Eq(\mathcal{T},\mathcal{M}_{\mathbb{M}},\mathcal{M}_{\mathbb{M}})  \;\;|\;\; \mathcal{T}\cup\mathcal{T} \;\;|\;\; \mathcal{T}\cap\mathcal{T} \;\;|\;\; \nonumber \\
&& \Type \;\;|\;\; \top_{\Type} \;\;|\;\; \top \;\;|\;\; \mathcal{J} \;\;|\;\; \mathcal{X}
\end{eqnarray}
\noindent These syntactic constructions represent base-level types, function types, dependent types, equality types, type unions and intersections, a base universe of small types, the union of all small types, the union of all types, judgments and execution states respectively.  Notice also that in Eq. \ref{eq:metta1}, $\mathcal{T}$ is defined by mutual recursion with the type  $\mathcal{M}_{\mathbb{M}}$, defined in Eq. \ref{eq:metta2}.  We then define $\mathcal{L}$ as $\mathcal{L} = \mathcal{S}\cup \mathcal{V}\cup \mathcal{K}\cup\mathcal{T} \times \mathbb{N}$.  Notice that $\mathcal{L}$ includes $\mathcal{T}$, so that types may simultaneously serve as labels.  Further, $\mathcal{S}=\{s_1,s_2,...\}$ and $\mathcal{V}=\{v_1,v_2,...\}$ denote collections of symbols and variables respectively, and $\mathcal{K}$ is a special set of $\mathbb{M}$ keywords/key-symbols:
\begin{eqnarray}
\mathcal{K} &=& \{:,\preceq,=,\rightarrow,\Eq,\funapp,\trans,@,\dagger\} 
\end{eqnarray}
\noindent Further, $\mathcal{L}$ includes an edge-specific identifier $\mathbb{N}$ to deduplicate edges which are identical in other respects.

The state of an $\mathbb{M}$ program is represented by a typed metagraph in the following space:
\begin{eqnarray}\label{eq:metta2}
\mathcal{M}_{\mathbb{M}}=\sum \preceq_T:(\mathcal{T}\times\mathcal{T}\rightarrow \mathbb{B}).\sum M:\mathcal{M}_{(\mathcal{T},\mathcal{L},\preceq_T)}.C_{\mathbb{M}}(M,\preceq_T)
\end{eqnarray}
\noindent Hence, this is the space of all metagraphs over $\mathcal{L}$ and $\mathcal{T}$, with a varying $\preceq_T$ relation, where $C_{\mathbb{M}}(M,\preceq_T)$ represents a set of '$\mathbb{M}$-specific constraints' on the structure of the metagraph (to be outlined below).  This state represents the {\em Atomspace} of the program, and the subgraphs of the Atomspace are the individual atoms (as in MeTTa, see \cite{potapov_21,goertzel_21}).  We note that, since $\mathbb{M}$ serves both as a language for defining programs and type-systems within which these programs are embedded, the atoms may represent base-level propositions and programs (expressions), as well as judgments and computational state information, as reflected by their types.  The $\mathbb{M}$-specific constraints, $C_{\mathbb{M}}(M,\preceq_T)$, determine the interaction of the keywords/key-symbols with the type system:
\begin{eqnarray}\label{eq:condits}
\forall m &\in& M.\exists n,n_1 : \mathbb{N}. \nonumber\\
m &=& \edge(2,\mathcal{J},(:,n),[\top_{\Type} \; \top])  \vee \nonumber\\
m &=& \edge(2,\mathcal{J},(\preceq,n),[\Type \; \Type]) \wedge \nonumber\\
&& (m_M[1]= \edge(0,\Type,(t_{n_1},0),[]) \wedge \nonumber\\
&& m_M[2]=\edge(0,\Type,(t_{n_2},0),[])\Rightarrow(t_{n_1}\preceq t_{n_2})) \vee \nonumber\\
m &=& \edge(2,\mathcal{J},(=,n),[\top_{\Type} \; \top_{\Type}]) \vee \nonumber\\
m &=& \edge(2,\Type,(\rightarrow,n),[\Type \; \Type]) \wedge \nonumber\\
&& ((m_1)_M[2] = m \wedge l(m_1)=(:,n_1) \wedge t(m_M[1])=A \wedge \nonumber\\
&& t(m_M[2])=B  \Rightarrow t((m_1)_M[1])\preceq A\rightarrow B) \vee \nonumber\\
m &=& \edge(2,\Type,(\rightarrow,n),[\Type \; \top]) \wedge \nonumber\\
&& ((m_1)_M[2] = m \wedge l(m_1)=(:,n_1) \wedge t(m_M[1])=A \wedge \nonumber\\
&& m_M[2]=m_2  \Rightarrow t((m_1)_M[1])\preceq \prod a:A.m_2) \vee \nonumber\\
m &=& \edge(3,\Type,(\Eq,n),[\Type  \; t_{n_1} \: t_{n_1}]) \wedge \nonumber\\
&& m_M[1] = \edge(0,\Type,(t_{n_1},0),[]) \wedge \nonumber\\
&& ((m_1)_M[2] = m \wedge l(m_1)=(\Eq,n_1) \wedge l(m_M[1])=(T,n_1) \wedge \nonumber\\
&& m_M[2]=A \wedge m_M[3]=B  \Rightarrow t((m_1)_M[1])=\Eq(T,A,B)) \vee \nonumber\\
m &=& \edge(2,\top,(\trans,n),[\top \; \top]) \vee \nonumber\\
m &=& \edge(1,\mathcal{X},(@,n),[\top]) \wedge \nonumber\\
m &=& \edge(1,\mathcal{X},(\dagger,0),[\top]) \wedge \nonumber\\
m &=& \edge(0,\top,(\mathcal{S}\cup \mathcal{V}\cup\mathcal{T},n),[]) \wedge \nonumber\\
m &=& \edge(2,B,(\funapp,n),[A\rightarrow B' \; A]) \wedge B'\preceq B \vee \nonumber\\
m &=& \edge(2,B,(\funapp,n),[\prod a:A.m_1 \; A]) \wedge \nonumber\\
&& m_1[a=m_M[1]]\preceq B \vee \nonumber\\
m &=& \connect(\_,\_,\_,\_,\_) \wedge \nonumber\\
\forall n,n_1,n_2&:&\mathbb{N}. t_n \preceq  \top \wedge \nonumber\\
s_n &: &  \top_{\Type} \vee s_n:\Type \wedge \nonumber\\
v_n &: & \top_{\Type} \vee v_n:\Type \wedge \nonumber\\
t_n &: & \Type \wedge  \nonumber\\
(t_{n_1} &\preceq & t_{n_2} \wedge  t_{n_2} \preceq  t_{n} \Rightarrow  t_{n_1} \preceq  t_{n}) \wedge  \nonumber\\
t_n &\preceq & t_n \cup t_{n_1} \wedge  \nonumber\\
t_n \cap t_{n_1} &\preceq & t_n
\end{eqnarray}
\noindent where the notation $m_M[n]$ denotes the $n$'th target of subgraph $m$ in metagraph $M$, $t[m]$ and $l[m]$ denote the type and label of metagraph $m$ respectively, and we write $a:A$ as shorthand for 'there exits an $:$-edge in $M$ connecting $a$ and $A$'.  We note that, for convenience, the above formulation does not include some constructions that may be appropriate in a full implementation, but can be derived from others.  For instance, tuples can be constructed by introducing a dependent function $\tuple : \prod A,B:\Type. A\rightarrow B \rightarrow \Type$.  The left and right projection functions are then defined by $\pi_1(\tuple(A,B,a,b))=a$ and $\pi_2(\tuple(A,B,a,b))=b$.  Dependent sums can likewise be defined as dependent tuples, $\tuple' : \prod A:\Type. \prod B:(A\rightarrow \Type).\prod a:A. B(a) \rightarrow \Type$.

\subsection{Labeled transition system based on metagraph rewriting}\label{sec:3a}

In guarded cubical type theory, a guarded labeled transition system (GLTS) may be defined via a state-space $X$, a space of actions $A$, and a function mapping states to sets of (action,state) pairs, $f:X\rightarrow P_\text{fin}(A\times \triangleright X))$, where $P_\text{fin}$ is the finite powerset functor.  The space of all processes, or runs of the GLTS may the be defined as the final coalgebra of the following functor: $\text{Proc}=\fix X.P_\text{fin}(A\times \triangleright(\alpha : \mathbb{T}).X[\alpha]))$ (see \cite{mogelborg_19}).  In order to characterize the process of evaluation in $\mathbb{M}$, we characterize the computational dynamics of $\mathbb{M}$ via a GLTS.  Here, the state space is the space of all $\mathbb{M}$ metgraphs, $X=\mathcal{M}_{\mathbb{M}}$.  The actions are specified by single pushout (SPO) rewriting rules, or sequences of such rules.  We therefore introduce the type, $\mathcal{A}'=\mathcal{M}^{(L,R)}_{\mathbb{M}}\times \hom_p(\mathcal{M}_{\mathbb{M}})$, whose values $(M',\phi)$ consist of a $\mathbb{M}$ metagraph whose label set is  $\mathcal{L}'=\mathcal{L}\times\{L,R, LR\}\times\{[],*,**\}$, i.e. identical to above, but with $L$ and $R$ labels added to each edge to indicate its membership of the left or right-hand side of the rule (notice that these may overlap), * and ** to indicate the input and output nodes of the rule (see below), and $\phi$, a partial metagraph homomorphism between the $L$ and $R$ metagraphs of $M'$ (defining a partial metagraph homomorphism as in \cite{goertzel_21}).  Since we wish to allow sequences of rewrite rules as actions, we define the full action space to be $\mathcal{A}=\sum n:\mathbb{N}.\Vect(n,\mathcal{A}')$,  and write the members of $\mathcal{A}$ as $a_1\circ a_2 \circ ... \circ a_n$, where $a_{1...n}:\mathcal{A}'$.  The dynamics are then defined (via $f$) by mapping a given metagraph state $M_1$ to the set of all pairs $(A,M_2)$ such that $M_2$ results from an application of action $A$ to $M_1$.  For individual rewrite rules $a\in\mathcal{A}'$, their action is determined via a partial homomorphism between $a$ and $M_1$.  We note that, when there are no partial homomorphisms between $a$ and $M_1$, or when the rewrite rule produces an invalid $\mathbb{M}$ graph, we set $M_2=M_1$.  Further, we note that the update may change the $\preceq$ relation, for instance by introducing an edge of the form $t_1\preceq t_2$.

\subsection{$\mathbb{M}$-interpretation as metagraph dynamics}\label{sec:3b}

\begin{figure*}[!t]
\centering
\includegraphics[width=4.75in]{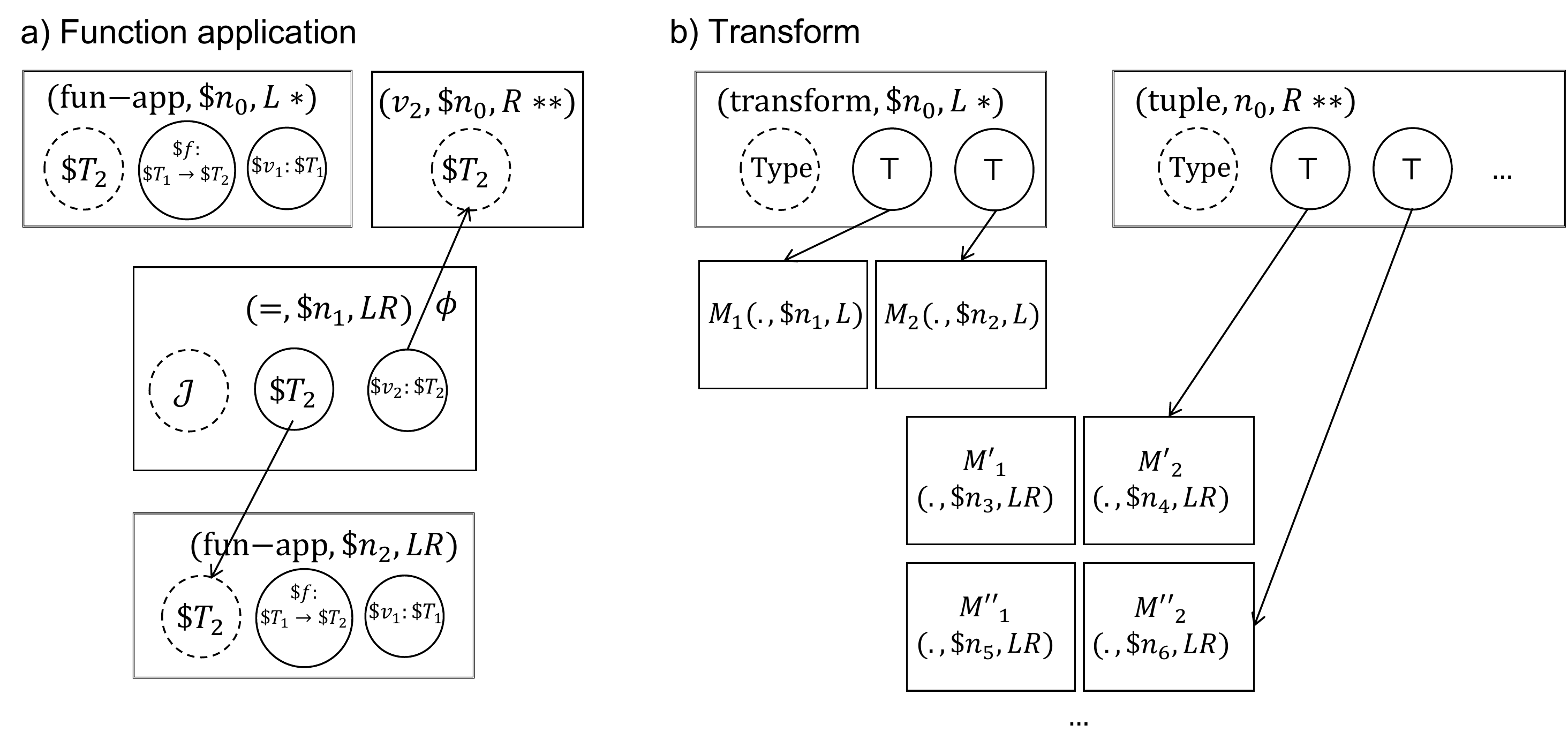}
\caption{Metagraph rewriting rules.  Notation as in Fig. 1.  Subgraphs involving only one variable are not shown explicitly, but notated directly on the targets they are connected to.  See Eqs. \ref{eq:rw1} and \ref{eq:rw2} for explicit expressions for the graphs.}
\label{fig:fig2}
\end{figure*}

We can now describe interpretation in $\mathbb{M}$ via the GLTS defined above.  To do so, we map specific symbols/edges in the metagraph to actions in $\mathcal{A}$ (corresponding to the grounding domain $F$ in \cite{goertzel_21}).  Specifically, edges carrying symbols of a function type, $A\rightarrow B$, dependent product type, $\prod a:A.B$, or the $\trans$ symbol, are mapped to specific forms of rewrite rule, as specified below.  All other edges are mapped to the $\nul$ transform.  Fig. \ref{fig:fig2} specifies the general forms of the rewrite rules for function application, and transform rules (we note the $\trans$ is  equivalent to the 2-argument $\match$ keyword/function in the current version of the MeTTa language, see \cite{trueagi_21}).  The dependent product rule is identical to Fig. \ref{fig:fig2}a, with $A\rightarrow B$ replaced with $\prod a:A.m_1$  For explicitness, we give these below also in equational form.  We note that, for convenience variable names are denoted using $\$$, although these should be ultimately mapped to the names $v_1, v_2,...$.
\begin{eqnarray}\label{eq:rw1}
R^1_{\funapp} &=& \edge(2,\$T_2,(\funapp,\$n_0,L),[\$T_1\rightarrow\$T_2 \; \$T_1]) \nonumber \\
R^2_{\funapp} &=& \edge(2,\mathcal{J},(=,\$n_1,LR),[\$T_2 \; \$T_2]) \nonumber \\
R^3_{\funapp} &=& \edge(2,\$T_2,(\funapp,\$n_2,LR),[\$T_1\rightarrow\$T_2 \; \$T_1]) \nonumber \\
R^4_{\funapp} &=& \edge(0,\$T_1\rightarrow\$T_2,(\$f,\$n_3,LR),[]) \nonumber \\
R^5_{\funapp} &=& \edge(0,\$T_1,(\$v_1,\$n_4,LR),[]) \nonumber \\
R^6_{\funapp} &=& \edge(0,\$T_2,(\$v_2,\$n_5,LR**),[]) \nonumber \\
R^7_{\funapp} &=& \connect(\connect(R^1_{\funapp},R^4_{\funapp},\top,\nul,\{(1,0)\})), \nonumber \\
&& R^5_{\funapp},\top,(\nul,\nul,*),\{(5,0)\}) \nonumber \\
R^8_{\funapp} &=& \connect(\connect(R^3_{\funapp},R^4_{\funapp},\top,\nul,\{(1,0)\})),\nonumber \\
&&R^5_{\funapp},\top,\nul,\{(5,0)\}) \nonumber \\
R^9_{\funapp} &=& \connect(\connect(R^2_{\funapp},R^3_{\funapp},\top,\nul,\{(1,0)\})),\nonumber \\
&&R^5_{\funapp},\top,\nul,\{(5,0)\}) \nonumber \\
R_{\funapp} &=& R^7_{\funapp} \cup R^8_{\funapp} \cup R^9_{\funapp} 
\end{eqnarray}
\begin{eqnarray}\label{eq:rw2}
R^1_{\trans} &=& \edge(2,\Type,(\trans,\$n_0,L),[\top \; \top]) \nonumber \\
R^2_{\trans} &=& \connect(\connect(R^1_{\trans},\$M_1,\top,\nul,\{(1,0)\})),\nonumber \\
&&\$M_2,\top,(\nul,\nul,L*),\{(5,0)\}) \nonumber \\
R^3_{\trans} &=& \edge(2,\Type,(\tuple,\$n_0,R**),[\top \; \top]) \nonumber \\
R^4_{\trans} &=& \$M'_1 \cup  \$M''_1 \cup  \$M'_2 \cup  \$M''_2 \cup \edge(0,\top,(\nul,\nul,LR),[]) \nonumber \\
R^5_{\trans} &=& \connect(R^3_{\trans},R^4_{\trans},\top,(\nul,\nul,\nul),\{(1,1),(2,2)\}))\nonumber \\
R_{\trans} &=& R^2_{\trans} \cup R^5_{\trans} 
\end{eqnarray}
\noindent In Eq. \ref{eq:rw2}, $M'_1$ and $M'_2$ denote metagraphs isomorphic to $M_1$ and $M_2$, using a disjoint set of variables, while $M''_1$ and $M''_2$ are defined similarly, with variables disjoint to the previous subsets.  The rule in Eq. \ref{eq:rw2} is defined so as to return a 2-tuple of matches; in general, the size of the tuple returned should be large enough to allow for any number of matches (i.e the number of nodes in $M$), and if the number of matches is less than this, it will be padded with $\nul$ values.

\vspace{0.25cm}
\noindent \textbf{fun-app nodes.} For a given annotated $\funapp$ node, i.e. $\connect(@,F,\nul,\{(1,0)\}))$, where $@=\edge(1,\mathcal{X},(@,n),[\top])$ and $F$ is a graph consisting of a target $\funapp$ node and its two arguments, the full rewrite rule $\rewrite_{F}$ is found by forming a metagraph homomorphism between  $R^7_{\funapp}$ (labeled by $*$ as the input of the rule), and $F$, replacing the variables in $R_{\funapp}$ by their values in $F$.  The resulting graph is denoted $R_{\funapp}(F)$.  The rule $\rewrite_{F}$ is then defined by the subgraphs $L=\connect(\$M_0,\connect(@,l_{\funapp}(F),\nul,\{(1,0)\})),\nul,\{((n_1,m_1),(n_2,m_1),...\})$, $R=\connect((\$M_0,r_{\funapp}(F),\nul,\{((n_1,m),(n_2,m),...\})$, where $M_0$ is the graph of all nodes in $M$ targeting $F$, $m_1$ is the index of the $\funapp$ node in $F$, $m_2$ is the index of the $**$ output node in $r_{\funapp}(F)$, and $\phi$ is defined by the partial homomorphism consisting of the identity map on all nodes labeled $LR$.

\vspace{0.25cm}
\noindent \textbf{transform nodes.} For a given annotated $\trans$ node, the full rewrite rule is defined similarly.  Hence, for $\connect(@,F,\nul,\{(1,0)\}))$, where $@=\edge(1,\mathcal{X},(@,n),[\top])$ and $F$ is a graph consisting of a target $\trans$ node and its two arguments, the full rewrite rule $\rewrite_{F}$ is found by forming a metagraph homomorphism between $R^2_{\trans}$ (labeled by $*$ as the input of the rule), and $F$, replacing the variables in $R_{\trans}$ by their values in $F$.  The resulting graph is denoted $R_{\trans}(F)$.  The rule $\rewrite_{F}$ is then defined by the subgraphs $L=\connect(\$M_0,\connect(@,l_{\trans}(F),\allowbreak \nul,\{(1,0)\})),\nul,\{((n_1,m_1),(n_2,m_1),...\})$, $R=\connect((\$M_0,r_{\trans}(F),\allowbreak \nul,\{((n_1,m),(n_2,m),...\})$, where $M_0$ is the graph of all nodes in $M$ targeting $F$, $m_1$ is the index of the $\trans$ node in $F$, $m_2$ is the index of the $**$ output node in $r_{\trans}(F)$, and $\phi$ is defined by the partial homomorphism consisting of the identity map on all nodes labeled $LR$.

\vspace{0.25cm}
\noindent \textbf{$\mathbb{M}$-evaluation.} The above provides groundings for activated nodes in a metagraph; as noted, nodes not of the form above result in a $\nul$ update.  Evaluation in $\mathbb{M}$ involves repeatedly updating the current pointed metgraph according to the grounding of the node currently pointed to.  The conditions in Eq. \ref{eq:condits} imply there will be at most one edge labeled with $\dagger$ in a metagraph, whose target $F$ specifies the rule by which the graph is updated.  This is expressed via the single partial function, $\update : \mathcal{M}_{\mathbb{M}} \rightarrow \mathcal{M}_{\mathbb{M}}$.  The action of $\update$ is determined by the form of $F$.  If $F$ is not an activated subgraph, i.e. it is not the target of an $@$-edge, the action $\update$ cannot be applied (i.e. evaluation halts).  If however $F$ is the target of an $@$-edge, $\update$ first checks if $F$ itself has any activated targets.  If so, then $\update$ simply applies a graph rewrite which moves the pointer $\dagger$ to the first such activated target (in the ordering of the edge).  If not, $\update$ applies $\rewrite_{F}$, which automatically ensures that the update will finish with $\dagger$ pointing to the output subgraph, labeled $**$.  These dynamics define a reduced GLTS, with $X=\mathcal{M}_{\mathbb{M}}$, $A=\{\update\}$, and $f(M)=\{(\update,M'|\update(M)=M')\}$.  Note that there may be multiple $M'$'s for which $\update(M)=M'$ if $\rewrite_{F}$ for a $\funapp$ node is non-deterministic.  Processes are defined by the fixed point $\text{Proc} =  \nu(P_\text{fin}(A\times \triangleright X)))$.  Normal forms of $\mathcal{M}_{\mathbb{M}}$ are metagraphs for which $\update$ cannot be applied (i.e. their grounding is $\nul$).  Processes which reach a normal form are said to be terminating, and the initial expression of the process is said to evaluate to the normal form reached.  Alternatively, certain expressions may not reach a normal form, resulting instead in a non-terminating computation.

\section{Bisimulation of type systems in $\mathbb{M}$}\label{sec:4}

As described in \cite{mogelborg_19}, in guarded cubical type theory, a bisimulation $R:X\rightarrow X \rightarrow U$ for the GLTS $(X,A,f)$ may be defined via the following dependent type:
\begin{eqnarray}\label{eq:bisim}
\isGLTSBisim_f R = \prod x,y:X.R(x,y) &\rightarrow& \nonumber \\
(\prod x':\triangleright X.\prod a:A.(a,x')\in f(x) &\rightarrow& \exists y':\triangleright X.\prod a:A.\nonumber \\
(a,y')\in f(y) &\times& \triangleright(\alpha:\mathbb{T}).R(x'[\alpha])(y'[\alpha]))\times  \nonumber \\
(\prod y':\triangleright X.\prod a:A.(a,y')\in f(y) &\rightarrow& \exists x':\triangleright X.\prod a:A.\nonumber \\
(a,x')\in f(x) &\times& \triangleright(\alpha:\mathbb{T}).R(x'[\alpha])(y'[\alpha])).
\end{eqnarray}
As shown in \cite{mogelborg_19}, this type is equivalent to the path type over the recursive data type of processes defined by the GLTS, $\text{Proc}=\fix X.P_\text{fin}(A\times \triangleright(\alpha : \mathbb{T}).X[\alpha]))$.  We may further define a bisimulation $R_2:X_1\rightarrow X_2 \rightarrow U$ between two GLTS's over a common action space, $(X_1,A,f_1)$ and $(X_2,A,f_2)$ via a bisimulation over their coproduct (see \cite{baier_08}):
\begin{eqnarray}\label{eq:bisim2}
\is2GLTSBisim_{f_1,f_2} R_2 &=& \isGLTSBisim_{f_1+f_2} R'_2 \times \forall x_1:X_1. \exists x_2:X_2.R_2(x_1,x_2) \times \nonumber \\
&& \forall x_2:X_2. \exists x_1:X_2.R_1(x_1,x_2)
\end{eqnarray}
where $R'_2 : (X_1+X_2) \rightarrow (X_1+X_2) \rightarrow U$, $R'_2((a,x),(b,y))=R(x,y)$ when $a=1\wedge b=2$, $R'_2(x,y)=\bot$ otherwise, and $f_1+f_2:(X_1+X_2)\rightarrow \mathcal{P}(A\times (X_1+X_2))$ defined similarly.  Since $R_2(x_1,x_2)$ contains at least one matching element for each $x_1$ and $x_2$, we may extract functions $g_1:X_1\rightarrow X_2$ and $g_2:X_2\rightarrow X_1$ as subsets of $R_2$, where an element in the codomain of each is chosen arbitrarily when there are multiple matches in $R_2$.  Since bisimulation corresponds to path-equivalence for elements of each type, $g_1$ and $g_2$, we can choose $\pi_1$ and $\pi_2$ such that $g_1 \circ g_2 \circ \pi_1 = i_1$ and $g_ 2\circ g_1 \circ \pi_2 = i_2$, where $i_1$ and $i_2$ are the identity on $X_1$ and $X_2$ respectively, and $\pi_1(x)=x' \Rightarrow \exists p:\Path_{X_1}(x,x')$, $\pi_2(x)=x' \Rightarrow \exists p:\Path_{X_2}(x,x')$.  Hence, $(g_1,g_2)$ is an equivalence between the recursive process types $\text{Proc}_1$ and $\text{Proc}_2$ of the two GLTS's, meaning that $\Path_{U}(\text{Proc}_1,\text{Proc}_2)$ is inhabited by univalence.

For a given type system, its computational content may be modeled by a GLTS by setting $X$ to be the type of expressions in the system, $A$ to contain an $\update$ action along with `actions' corresponding to the judgmental and syntactic relations between expressions (e.g. is-of-type, is-of-subtype, is-a-body-of-lambda-term, and their opposite relations), and $f$ to be the relation over expressions corresponding to the reduction relation in the system for the action $\update$ (for instance $\beta$-reduction).  To show that $\mathbb{M}$ can be used as a metalanguage for a given type system, we thus show that there is a bisimulation between $\mathbb{M}$ with a specific form of Atomspace (i.e. containing specific atoms and/or additional constraints to those of Eq. \ref{eq:condits}), along with an expanded action space to incorporate the typing and syntactic relations relevant to the specific system, and the GLTS corresponding to computation in the target type system; hence the process spaces induced by the two systems are equivalent.  Below, we sketch how this can be achieved for three type systems of interest, focusing on the how the computational dynamics of the $\update$ rule correspond to reduction in the target system (the typing and syntactic relations in each system straightforwardly correspond in $\mathbb{M}$ to the inbuilt typing relation and relationships definable in terms of submetagraph composition respectively).

\subsection{Simply typed lambda calculus}\label{sec:4a}

The syntax for the simply typed lambda calculus may be defined via mutually recursive definitions of variable, type and expression datatypes:
\begin{eqnarray}\label{eq:simple1}
\mathcal{V} &::=& v_n \nonumber \\
\mathcal{T} &::=& t_n \;\;|\;\; \mathcal{T} \rightarrow \mathcal{T} \nonumber \\
\mathcal{E} &::=& \mathcal{V} \;\;|\;\; (\mathcal{E} \; \mathcal{E}) \;\;|\;\; \lambda v_n:\mathcal{T}.\mathcal{E}
\end{eqnarray}
\noindent We refrain from explicitly stating the rules for type assignment as can be found in \cite{barendregt_92}, which determine a typing relation $\_:\_$ between $\mathcal{E}$ and $\mathcal{T}$ given a context $\Gamma$, which can be modeled as a partial map from $\mathcal{V}$ to $\mathcal{T}$.  Together, these determine a set of valid expressions, $\mathcal{E}_{(\_:\_,\Gamma)}$, and the computational dynamics is defined by the $\beta$-reduction relation over this type:
\begin{eqnarray}\label{eq:simple2}
((\lambda v_{n_1}:t_{n_2}.e_{n_3}) \; e_{n_4}) \rightarrow_{\beta} e_{n_3}[v_{n_1}/e_{n_4}]
\end{eqnarray}
\noindent where $a[b/c]$ denotes substitution of $b$ for $c$ in $a$, where any bound variables in $c$ are renamed so as not to clash with bound variables in $a$.

To simulate the simply typed lambda calculus in $\mathbb{M}$, we restrict the $\mathbb{M}$ atomspace to include only metagraphs labeled with types using the restricted type syntax of Eq. \ref{eq:simple1}, and including only keywords/symbols $\{:,=,\rightarrow,\funapp,@,\dagger\}$.  Then, we add the following constraint to those of Eq. \ref{eq:condits}:
\begin{eqnarray}\label{eq:simple3}
\forall m \in M. l(m)=(:,n_1) \Rightarrow (m_M[1]\in \mathcal{S} \vee \mathcal{V}) \wedge m_M[2]\in \mathcal{T}
\end{eqnarray}
\noindent Hence, all typing relations are between symbols or variables (representing global and local variables respectively) and types.  The context $\Gamma$ is then represented by an atomspace consisting of a set of $:$ edges between symbols and types.  A given lambda expression $e=\lambda x:t_1.e'$, where $e':t_2$ is then simulated by choosing an unused symbol, $f_e\in \mathcal{S}$, and introducing the following atoms to atomspace:
\begin{eqnarray}\label{eq:lambda}
&& (: \; f_e \; (\rightarrow \; t_1 \; t_2)) \nonumber \\
&& (= \; (f_e \; \$x) \; m_{e'})
\end{eqnarray}
\noindent where $m_{e'}$ is the metagraph corresponding to expression $e'$ (we note that Eq. \ref{eq:lambda} defines a combinator corresponding to the lambda term $e$).  With the atomspace so specified, reduction of an expression $e$ in context $\Gamma$ in the simply typed lambda calculus corresponds to repeated application of $\update$ to the pointed atomspace containing $\Gamma$ and $m_e$, with $@$ edges attached to all function application nodes, and the $\dagger$ pointing to $m_e$.  The computation terminates with $\dagger$ pointing to the normal form of $e$.  The required bisimulation thus involves pairing tuples $(\Gamma,e)$ in the simply typed lambda calculus with their corresponding pointed atomspaces in $\mathbb{M}$.  We note further that the untyped lambda calculus can be defined by simply removing $\mathcal{T}$ from the syntax in Eq. \ref{eq:simple1}, and letting lambda expressions take the form $\lambda v_n.\mathcal{E}$.  All members of $\mathcal{E}$. are considered legal expressions, and the $\mathbb{M}$ bisimulation is achieved by converting all type symbols to $\top_{\Type}$, hence treating $\top_{\Type}$ as a Scott domain.

\subsection{Pure Type Systems}\label{sec:4b}

In a pure type system (PTS, \cite{barendregt_92}), types and terms are not distinguished syntactically.  PTS expressions follow the syntax:
\begin{eqnarray}\label{eq:pts1}
\mathcal{V} &::=& v_n \nonumber \\
\mathcal{C} &::=& s_{1...N} \nonumber \\
\mathcal{E} &::=& \mathcal{V} \;\;|\;\; \mathcal{C} \;\;|\;\; (\mathcal{E} \; \mathcal{E}) \;\;|\;\; \lambda v_n:\mathcal{E}.\mathcal{E} \;\;|\;\; \prod v_n:\mathcal{E}.\mathcal{E}
\end{eqnarray}
\noindent Here, $\mathcal{C}$ is a set of constant symbols, which in a PTS are used to represent {\em sorts}.  The typing relation $:$ for a PTS is defined via a set of axioms and rules.  The former consist of a set of judgements $\mathcal{A}=\{s_m:s_n|(m,n)\in A \subset N\times N\}$, and the latter a set of triplets $\mathcal{R}=\{(s_l,s_m,s_n)|(l,m,n)\in R \subset N\times N\times N\}$.  The typing rules for a PTS are identical to the typed lambda calculus, except for the introduction rule for dependent products, which takes the form:
\begin{prooftree}
\AxiomC{$\Gamma \vdash A:s_l \;\; \Gamma, A:s_l \vdash B:s_m \;\; (s_l,s_m,s_n)\in\mathcal{R}$}
\UnaryInfC{$\Gamma \vdash (\prod x:A.B):s_n$}
\end{prooftree}
\noindent The legal expressions then consist of the sorts, and any expression that can be typed in a context $\Gamma$, consisting of multiple typing judgments $e_1:e_2$.  The $\beta$-reduction relation is established identically to the simple lambda calculus above.  Notice that there is no restriction on the form of $\mathcal{A}$ and  $\mathcal{R}$; hence the typing relation $:$ may be arbitrary between sorts (and hence may contain cycles), while the dependent product (i.e. dependent function types) may live in arbitrary sorts with respect to their inputs.

To simulate a PTS in $\mathbb{M}$, we select a collection of fixed types $t_1...t_N$ to represent the sorts.  We then add edges of the following forms to atomspace:
\begin{eqnarray}\label{eq:pts1}
&&(: \; t_m \; t_n), \;\; \forall(s_m,s_n)\in\mathcal{A} \nonumber \\
&&(: \; (\rightarrow \; \$t_a \; \$t_b) \; (\trans \; (: \; \$t_a \; t_l)\wedge(: \; \$t_b \; t_m) \; t_n)), \;\; \forall(s_l,s_m,s_n)\in\mathcal{R}  \nonumber \\
&&(: \; (\prod \$x:\$t_a.\$m) \; (\trans \; (: \; \$t_a \; t_l)\wedge(: \; \$m \; t_m) \; t_n)), \;\; \forall(s_l,s_m,s_n)\in\mathcal{R} \nonumber \\
\end{eqnarray}
\noindent As above, lambda expressions are simulated by adding atoms of the form in Eq. \ref{eq:lambda} to the atomspace, and a context $\Gamma$ is simulated by adding atoms corresponding to the typing relations it contains.  Reduction of expression $e$ in context $\Gamma$ is simulated as previously by applying $\update$ to the pointed atomspace consisting of $\{\Gamma,e\}$ and the above constructions, along with $\dagger$ pointing to $e$.  Further, we note that we can use PTS's can be regarded as a type-theoretic analogue of non-well-founded sets; from this viewpoint, a cyclical $:$ relation corresponds to an accessible pointed graph (apg) underlying a non-well-founded set.  For instance, including the axiom $s_1:s_1$ in $\mathcal{A}$ defines $s_1$ as a type-theoretic analogue of a Quine atom.  We note, however, that in the type-theoretic context, a cyclic PTS carries more structure than a non-well-founded set, since the rules ($\mathcal{R}$) carry information about how the $\rightarrow$ constructor interacts with the $:$ relation.  An interesting conjecture though would be that appropriately defined PTS's provide bisimulations of systems of non-well-founded sets definable within a recursive datatype (via a coalgebra on the powerset functor, definable in GCTT), as a general system of set equations (\cite{barwise_96}) involving both $\in$ and $\rightarrow$ relations.

\subsection{Probabilistic dependent types}\label{sec:4c}

Finally, we outline a version of the probabilistic dependent type system introduced in \cite{warrell_18}, and its bisimulation in $\mathbb{M}$.  The syntax is a variation on the dependently typed lambda calculus:
\begin{eqnarray}\label{eq:pdts1}
\mathcal{V} &::=& v_n \nonumber \\
\mathcal{T} &::=& t_n \;\;|\;\; \prod v_n:\mathcal{T}.\mathcal{E} \;\;|\;\; \mathcal{D}(\mathcal{T}) \;\;|\;\; \mathcal{T}\cup\mathcal{T} \;\;|\;\; \mathcal{T}\cap\mathcal{T}  \;\;|\;\;  \Type \nonumber \\
\mathcal{E} &::=& \mathcal{V} \;\;|\;\; (\mathcal{E} \; \mathcal{E}) \;\;|\;\; \lambda v_n:\mathcal{T}.\mathcal{E} \;\;|\;\; \random_{\rho}(\mathcal{E},\mathcal{E})  \;\;|\;\;  \sample(\mathcal{E}) \;\;|\;\;  \thunk(\mathcal{E})  \nonumber \\
\end{eqnarray}
\noindent Further, we allow the judgments $\mathcal{E} : \mathcal{T}$ (typing),  $\mathcal{T} \preceq \mathcal{T}$ (subtyping), and $\mathcal{E} \rightarrow_{\beta}^{\rho} \mathcal{E}$ (weighted $\beta$-reduction), where $\rho\in\mathbb{R}$.  The typing rules are as for the dependent typed lambda calculus for expressions not involving subtypes or probabilistic terms.  The typing rules for subtypes include the standard $\Gamma \vdash a:A, \; A\preceq B \Rightarrow \Gamma \vdash a:B$, $\Gamma \vdash A,B:\Type \Rightarrow \Gamma \vdash A\cap B \preceq A, \; A\cap B \preceq B, \; A \preceq A\cup B \; B \preceq A\cup B$, $\Gamma \vdash A\preceq B \Rightarrow \prod v_n:B.\mathcal{E} \preceq \prod v_n:A.\mathcal{E}$, $\Gamma, x:t \vdash A\preceq B \Rightarrow \prod x:t.A \preceq \prod x:t.B$.  These interact with the probabilistic terms via the following special rules:
\begin{prooftree}
\AxiomC{$\Gamma \vdash a : t_1, \; b : t_2$}
\UnaryInfC{$\Gamma \vdash \random_{\rho}(a,b) : t_1\cup t_2$}
\end{prooftree}
\begin{prooftree}
\AxiomC{$\Gamma \vdash A : \Type,  p_A : \mathcal{D}(A)$}
\UnaryInfC{$\Gamma \vdash \sample(p_A) : A$}
\end{prooftree}
\begin{prooftree}
\AxiomC{$\Gamma \vdash a : A$}
\UnaryInfC{$\Gamma \vdash \thunk(a) : \mathcal{D}(A)$}
\end{prooftree}
\noindent where, we note that $\mathcal{D}(A)$ denotes the type of distributions over $A$ (so, for instance, if $a : t_1, \; b : t_2$, then $\thunk(\random_{\rho}(a,b)):\mathcal{D}(t_1\cup t_2)$).  For all expressions not involving probabilistic terms, $e_1 \rightarrow_{\beta} e_2$ in the dependent typed lambda calculus implies $e_1 \rightarrow_{\beta}^{1} e_2$ in the PDTS above.  For probabilistic terms, we have the following computational rules:
\begin{eqnarray}\label{eq:pdts2}
\random_{\rho}(a,b)  &\rightarrow_{\beta}^{\rho}& a \nonumber \\
\random_{\rho}(a,b)  &\rightarrow_{\beta}^{1-\rho}& a \nonumber \\
\sample(\thunk(p_A)) &\rightarrow_{\beta}^{1}& p_A 
\end{eqnarray}
Computationally, evaluation may proceed by stochastic $\beta$-reduction (i.e. sampling a reduction according to the weights $\rho$), or a 'full evaluation' may be made, by returning the set of all possible reduction sequences from a term, annotated with the total probability of each.  We note that in any given reduction sequence, $e_1 \rightarrow_{\beta}^{\rho} e_2$ for $\rho>0$ implies $t_2 \preceq t_1$ where $e_1:t_1,\; e_2:t_2$.

For the formulation in $\mathbb{M}$, we constrain the typing relation and encode lambda terms as in Eqs. \ref{eq:simple3} and \ref{eq:lambda}; further, as above we encode contexts $\Gamma$ by fixing atoms of the form $:$ in atomspace.  To encode the probabilistic terms, we choose fixed symbols $s_{1...4}$ to correspond to $\Distribution, \random, \sample, \thunk$.  Then, we fix the following atoms in atomspace:
\begin{eqnarray}\label{eq:pdts3}
&&(: \; \Distribution \; (\rightarrow \; \Type \; \Type)),  \nonumber \\
&&(: \; \random \; (\rightarrow \; \$t_1 \; \$t_2 \; \$t_1\cup\$t_2)),  \nonumber \\
&&(= \; (\random \; \$a \; \$b) \; \$a),  \nonumber \\
&&(= \; (\random \; \$a \; \$b) \; \$b),  \nonumber \\
&&(: \; \sample \; (\rightarrow \; (\Distribution \; \$t_1) \; \$t_1)),  \nonumber \\
&&(: \; \thunk \; (\rightarrow \; \$t_1 \; (\Distribution \; \$t_1))),  \nonumber \\
&&(= \; (\sample \; (\thunk \; \$a)) \; \$a)
\end{eqnarray}
\noindent Application of $\update$ to the pointed atomspace so defined, with $\dagger$ pointing to $m_e$ (corresponding to expression $e$), results in a simulation of a probabilistic reduction of $e$ in the PDTS above.  As defined, $\update$ will simulate the 'full evaluation' of all possible paths, and hence a bisimulation exists between full evaluation dynamics in the PDTS GLTS using $\beta\rho$-reduction and the GLTS defined by $\mathbb{M}$ with the restricted atomspace above.  We note that, in both cases, the weights on particular paths are lost, since the $\rho$ values are not explicitly recorded; however. it is straightforward to define a GLTS over the extended system, $(X\times \mathbb{R},A,f)$, where $f(x)=\{((x_1,p_1),a_1),((x_2,p_2),a_2),... \}$ denotes that action $a$ on $x$ results in $x_1$ with probability $p_1$, $x_2$ with probability $p_2$, and so on.

\section{Implementation of Bisimulation proof in a Guarded Cubical Type Theory type checker}\label{sec:4aa}

We briefly give an example to show the feasibility of our approach with an implementation of a bisimulation proof for a small-scale type system in a Guarded Cubical Type Theory type checker \cite{birkedal_16}.  Here, we model a minimal type system, which has one type constant $A:\Type$ with two constructors $v_1, v_2 :A$; one function constant $f_1:A\rightarrow A$, where $f_1(v_1)=v_2$ and $f_1(v_2)=v_1$; and includes the $\sample$ and $\thunk$ constructs, which are combined following the syntax of Eq. \ref{eq:pdts1}.  Our implementation models a fragment of this system where expressions are restricted to include at most three subexpressions.  Hence, valid expressions of the language include: $(f_1 \; (f_1 \; v_1))$, $(\thunk \; (f_1 \; v_2))$, $(\sample \; (\thunk \; v_1))$, $(f_1 \; v_2)$.  Our implementation in a Haskell-based Guarded Cubical Type Theory type checker \cite{birkedal_16} is given in Appendix A.  Here, we implement evaluation in this system via (i) a pattern matcher over an atomspace ('update'), and (ii) direct implementation of $\beta$-reduction via case analysis over the expression space ('beta3').  We define GLTS's using both forms of evaluation ('str1' and 'str2'), and finally derive a proof that these GLTS's are bisimilar ('bisim').  The code for this example is also provided at: \url{https://github.com/jwarrell/metta_bisimulation}

\section{Discussion}\label{sec:5}

In the above, we have introduced a formal meta-probabilistic programming language, formalized in GCTT, and proposed that bisimutations link the specific object-languages (or domain specific languages) outlined above with their simulations in $\mathbb{M}$.  Specifically, we have proposed that the restricted forms of $\mathbb{M}$ outlined in Secs. \ref{sec:4a} and \ref{sec:4b} and \ref{sec:4c} form bisimulations of the simply typed lambda calculus, arbitrary PTS's, and the target PDTS, respectively.  

\noindent Finally, we mention some of the areas of investigation opened up by the formal model outlined.  First, we note that, while we have focused on `full' probabilistic programming evaluation, other possibilities include investigation of sampling based evaluation which performs only one meta-graph update at each step, stochastically chosen from the possible graph rewriting locations.  Second, we intend to derive further bisimulations for other kinds of probabilistic logic, particularly, probabilistic paraconsistent logic \cite{goertzel_20b}, and probabilistic analogues of pure type systems \cite{barendregt_92}, which may be suitable for models involving infinite-order probabilities  \cite{goertzel_08}.  Lastly, we intend to expand our implementation of aspects of this framework in Guarded Cubical Agda  \cite{veltri_20} to provide more complete implementations of the metalanguage and type systems explored here.

\newpage

\section*{Appendices}

\appendix
\section{Proof of Bisimulation for Small-scale Type System in a Guarded Cubical Type Theory type checker}\label{app:A}

Below, we provide the code for the example discussed in Sec. \ref{sec:4aa}, which uses a Haskell-based GCTT type checker \cite{birkedal_16}.  The code for this example is also provided at: \url{https://github.com/jwarrell/metta_bisimulation}

\begin{figure*}
\includegraphics[width=4.75in]{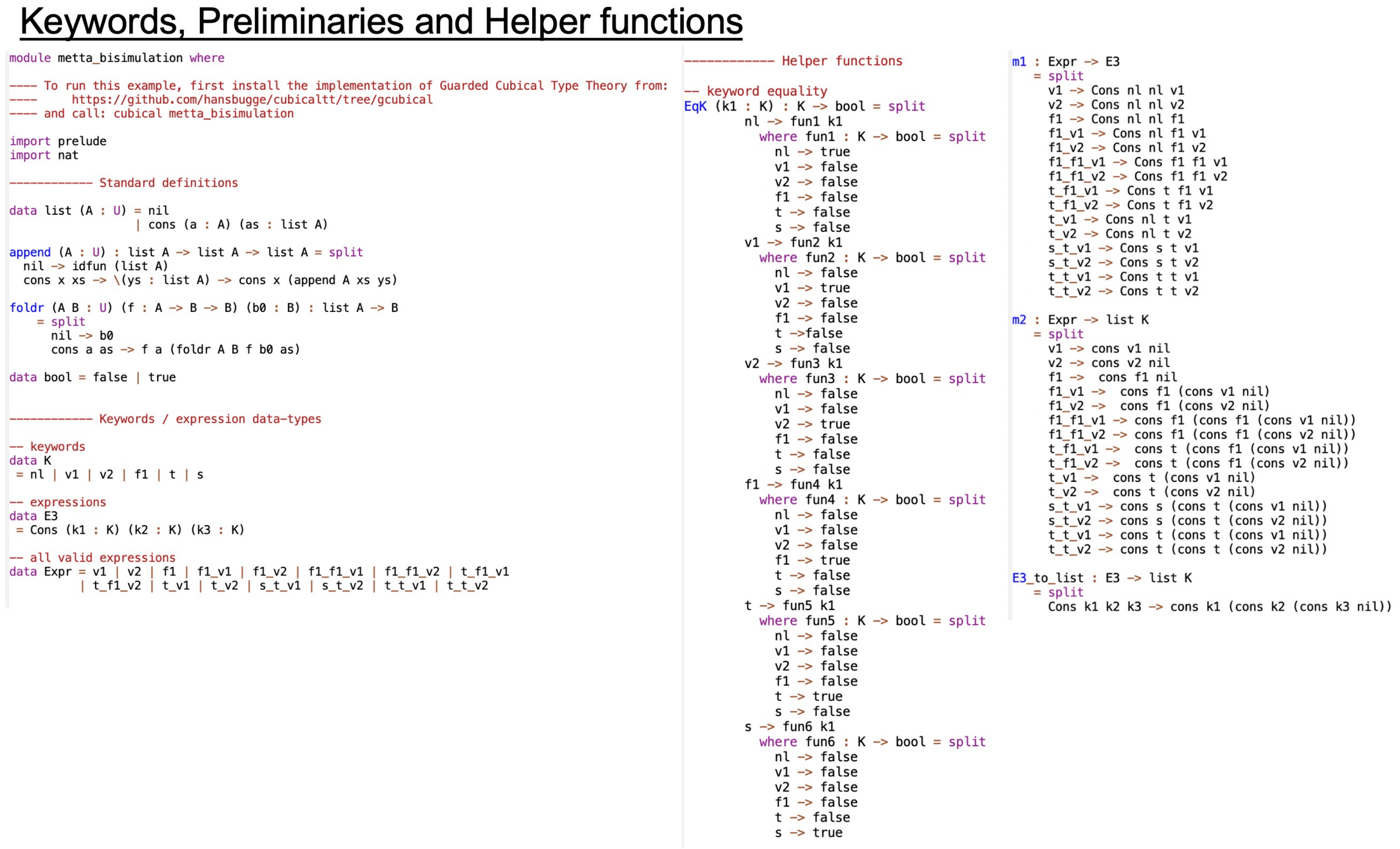}
\end{figure*}
\begin{figure*}
\begin{center}
\includegraphics[width=4.75in]{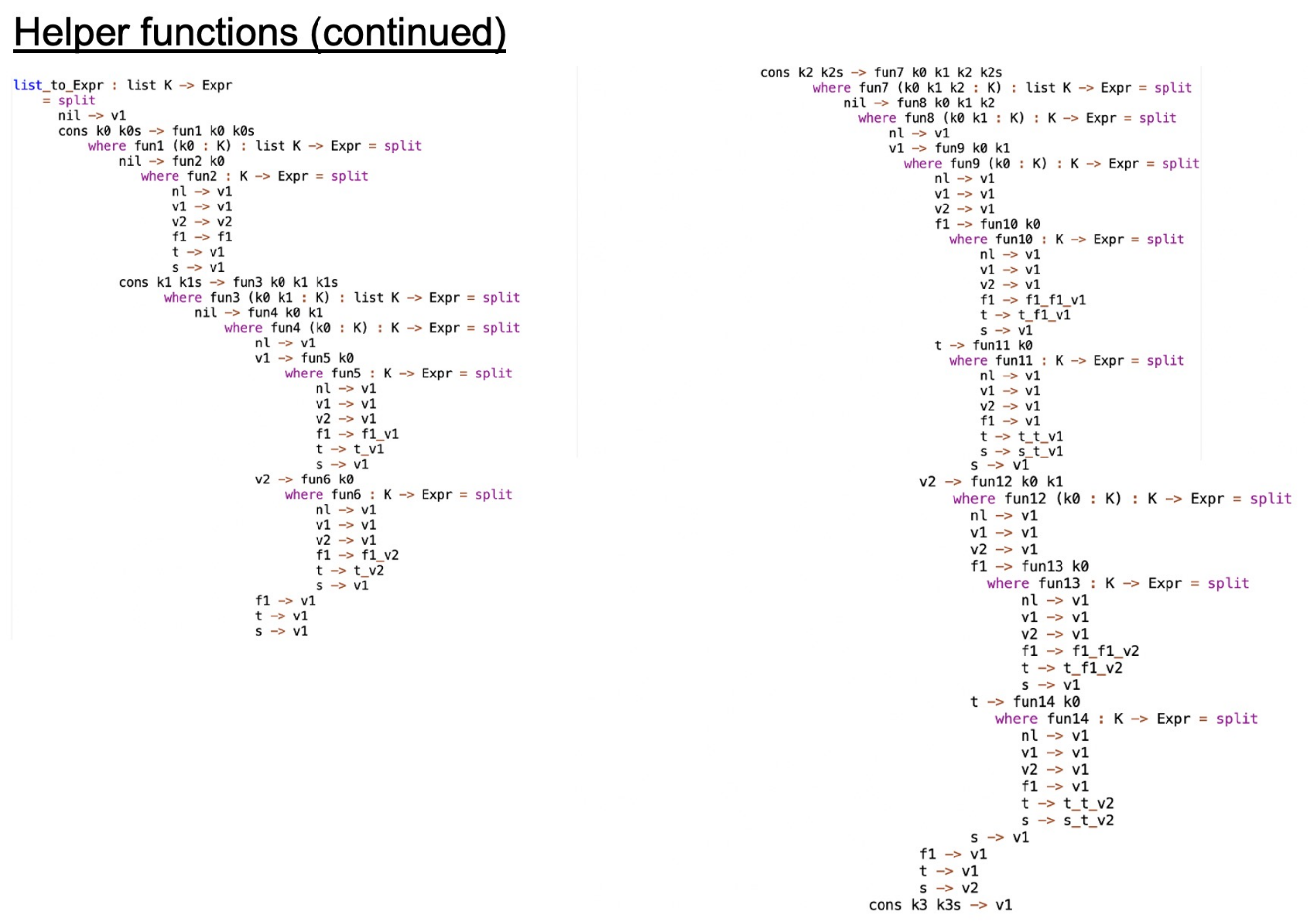}
\end{center}
\end{figure*}
\begin{figure*}
\includegraphics[width=4.75in]{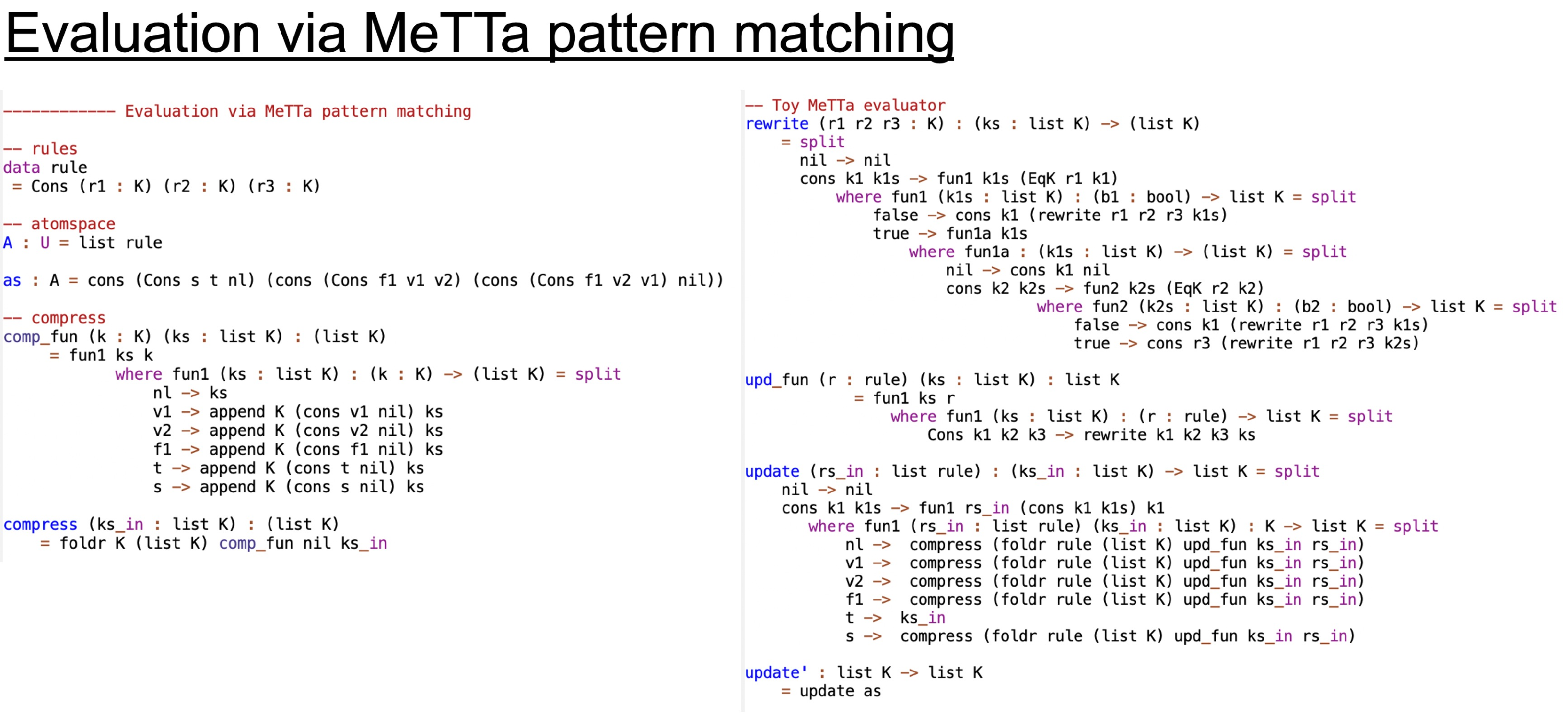}
\end{figure*}
\begin{figure*}
\includegraphics[width=4.75in]{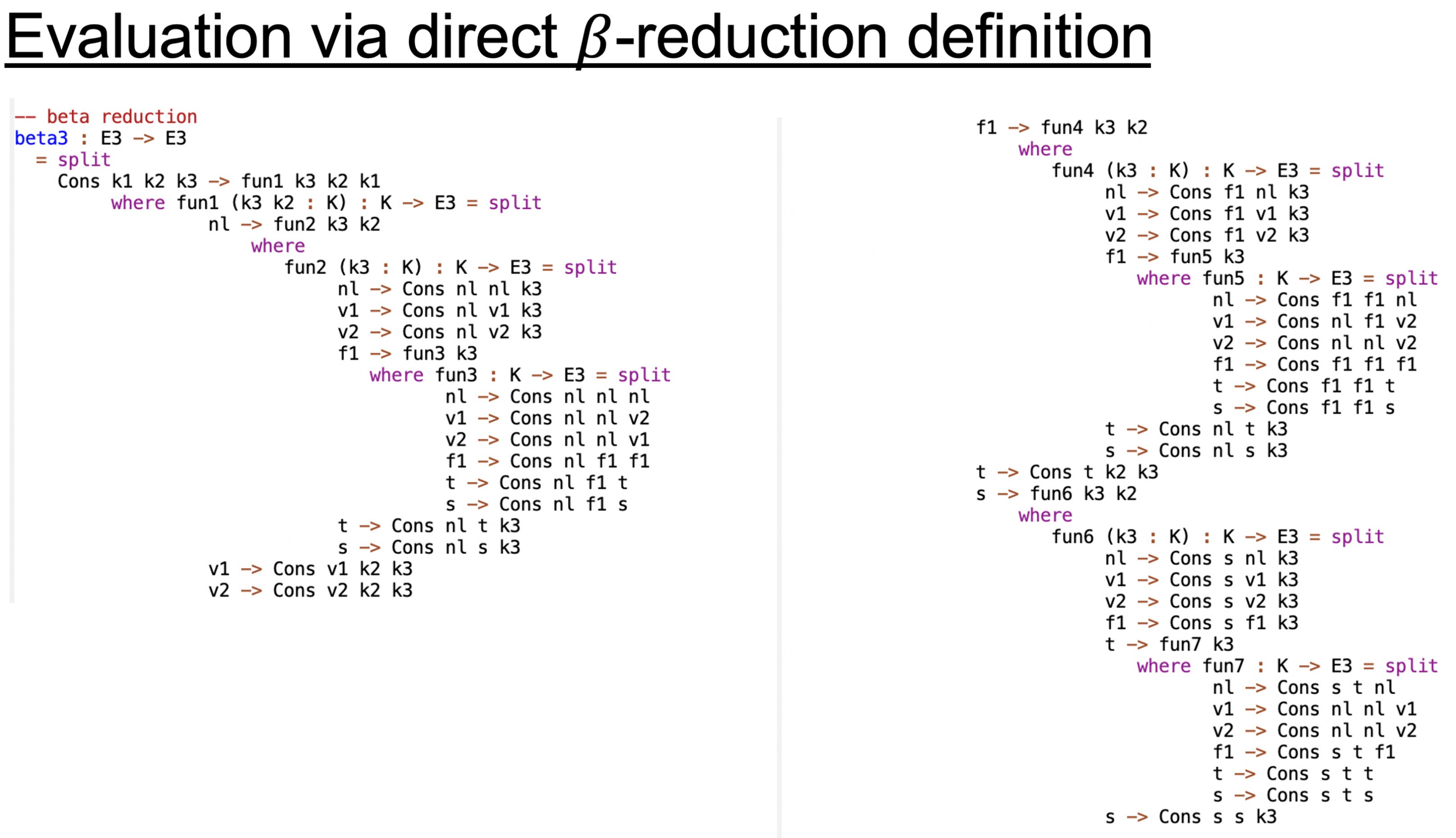}
\end{figure*}
\begin{figure*}
\includegraphics[width=4.75in]{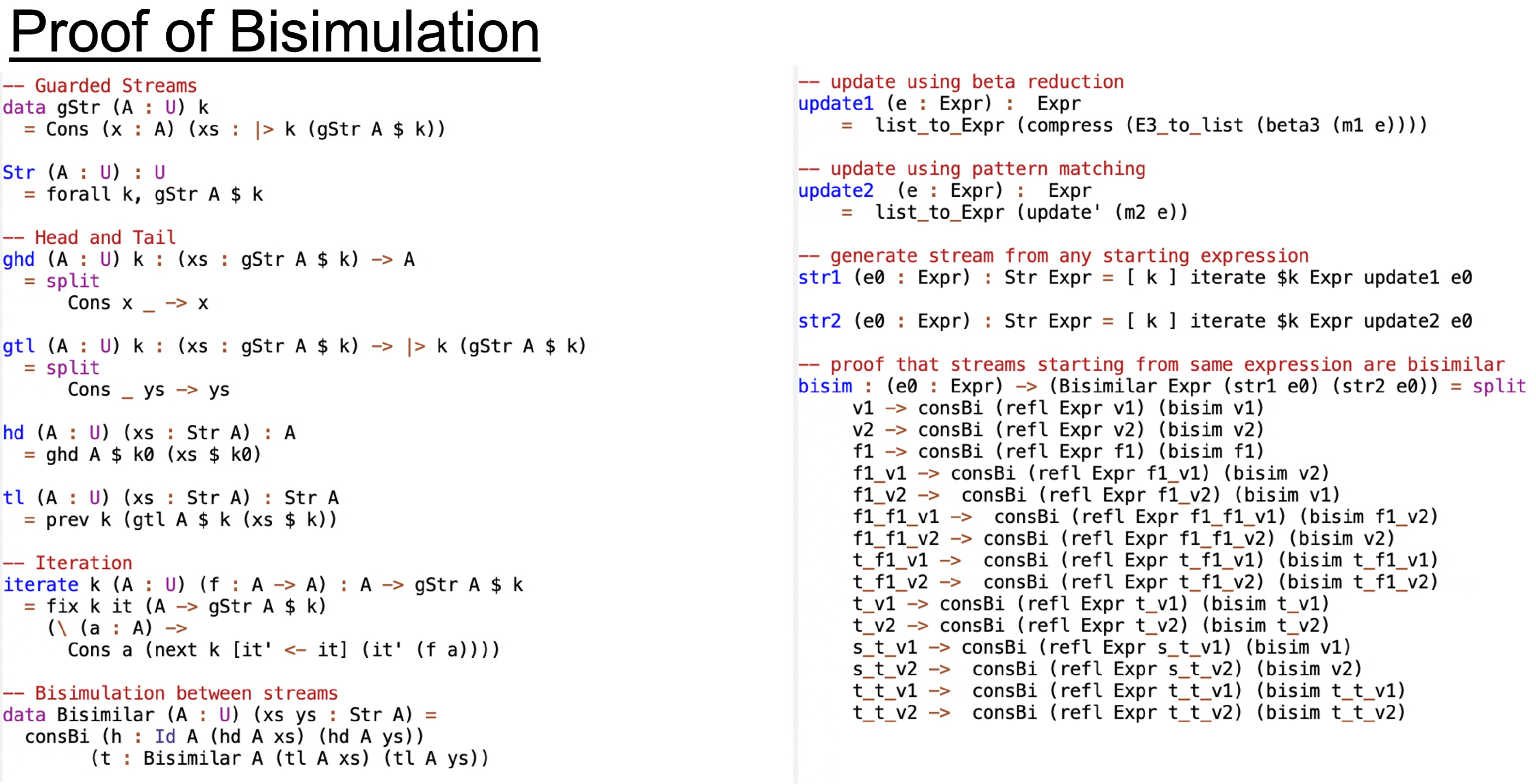}
\end{figure*}


\begin{thebibliography}{1}

\bibitem{baier_08}
Baier, C. and Katoen, J.P., 2008. Principles of model checking. MIT press.

\bibitem{barendregt_92}
Barendregt, Henk, and Lennart Augustsson, 1992. "Lambda Calculi with Types." Handbook of Logic in Computer Science 34: 239-250.

\bibitem{barwise_96}
Barwise, J. and Moss, L., 1996. Vicious circles: on the mathematics of non-wellfounded phenomena.

\bibitem{birkedal_16}
Birkedal, L., Bizjak, A., Clouston, R., Grathwohl, H.B., Spitters, B. and Vezzosi, A., 2016. Guarded cubical type theory: Path equality for guarded recursion. arXiv preprint arXiv:1606.05223.

\bibitem{goertzel_08}
Goertzel, B., 2008. Modeling Uncertain Self-Referential Semantics with Infinite-Order Probabilities.

\bibitem{goertzel_20}
Goertzel, B., 2020. Folding and Unfolding on Metagraphs. arXiv preprint arXiv:2012.01759.

\bibitem{goertzel_20b}
Goertzel, B., 2020. Paraconsistent Foundations for Probabilistic Reasoning, Programming and Concept Formation. arXiv preprint arXiv:2012.14474.

\bibitem{goertzel_21}
Goertzel, B., 2021. Reflective Metagraph Rewriting as a Foundation for an AGI `Language of Thought'. arXiv preprint arXiv:2112.08272.

\bibitem{harper_12}
Harper, R., 2012. Notes on logical frameworks. Lecture notes, Institute for Advanced Study, Nov, 29, p.34.

\bibitem{mogelborg_19}
Møgelberg, R.E. and Veltri, N., 2019. Bisimulation as path type for guarded recursive types. Proceedings of the ACM on Programming Languages, 3(POPL), pp.1-29.

\bibitem{mogelborg_19b}
Møgelberg, R.E. and Paviotti, M., 2019. Denotational semantics of recursive types in synthetic guarded domain theory. Mathematical Structures in Computer Science, 29(3), pp.465-510.

\bibitem{mokhov_17}
Mokhov, A., 2017. Algebraic graphs with class (functional pearl). ACM SIGPLAN Notices, 52(10), pp.2-13.

\bibitem{paviotti_15}
Paviotti, M., Møgelberg, R.E. and Birkedal, L., 2015. A model of PCF in guarded type theory. Electronic Notes in Theoretical Computer Science, 319, pp.333-349.

\bibitem{potapov_21}
Potapov, A., 2021. MeTTa language specification. \url{https://wiki.opencog.org/w/Hyperon}.

\bibitem{staton_16}
Staton, S., Wood, F., Yang, H., Heunen, C. and Kammar, O., 2016, July. Semantics for probabilistic programming: higher-order functions, continuous distributions, and soft constraints. In 2016 31st annual acm/ieee symposium on logic in computer science (lics) (pp. 1-10).

\bibitem{trueagi_21}
TrueAGI, 2021. Hyperon-experimental repository. \url{https://github.com/trueagi-io/hyperon-experimental}.

\bibitem{veltri_20}
Veltri, N. and Vezzosi, A., 2020, January. Formalizing $\pi$-calculus in guarded cubical Agda. In Proceedings of the 9th ACM SIGPLAN International Conference on Certified Programs and Proofs (pp. 270-283).

\bibitem{vezzosi_21}
Vezzosi, A., Mörtberg, A. and Abel, A., 2021. Cubical Agda: A dependently typed programming language with univalence and higher inductive types. Journal of Functional Programming, 31.

\bibitem{warrell_18}
Warrell, J. and Gerstein, M., 2018. Dependent Type Networks: A Probabilistic Logic via the Curry-Howard Correspondence in a System of Probabilistic Dependent Types. In Uncertainty in Artificial Intelligence, Workshop on Uncertainty in Deep Learning. \url{http://www. gatsby. ucl. ac. uk/~ balaji/udl-camera-ready/UDL-19. pdf}.

\end{thebibliography}
\end{document}